\documentclass[journal]{IEEEtran}

\usepackage{graphics}
\usepackage{multirow}
\usepackage{amsmath}
\usepackage{graphicx}
\usepackage{float}
\usepackage{siunitx}
\usepackage{array,booktabs,arydshln,xcolor}
\usepackage[caption=false]{subfig}
\usepackage{color}

\makeatletter
\newcommand{\rmnum}[1]{\expandafter\@slowromancap\romannumeral #1@}
\makeatother

\makeatletter
\def\normaljustify{%
  \let\\\@centercr\rightskip\z@skip \leftskip\z@skip%
  \parfillskip=0pt plus 1fil}
\makeatother

\begin{document}

\title{Utilizing Deep Learning Towards Multi-modal Bio-sensing and Vision-based Affective Computing}

\author{Siddharth~\IEEEmembership{Student Member,~IEEE},
        Tzyy-Ping Jung~\IEEEmembership{Fellow,~IEEE}, and Terrence J. Sejnowski~\IEEEmembership{Life Fellow,~IEEE}
\thanks{Siddharth is with the Department of Electrical and Computer Engineering, University of California San Diego, La Jolla, CA, 92093 USA, e-mail: ssiddhar@eng.ucsd.edu}
\thanks{Tzyy-Ping Jung and Terrence J. Sejnowski are with Institute for Neural Computation, University of California San Diego, La Jolla, CA, 92093 USA, e-mail: jung@sccn.ucsd.edu, terry@salk.edu}
}

\maketitle

\begin{abstract}
In recent years, the use of bio-sensing signals such as electroencephalogram (EEG), electrocardiogram (ECG), etc. have garnered interest towards applications in affective computing. The parallel trend of deep-learning has led to a huge leap in performance towards solving various vision-based research problems such as object detection. Yet, these advances in deep-learning have not adequately translated into bio-sensing research. This work applies novel deep-learning-based methods to various bio-sensing and video data of four publicly available multi-modal emotion datasets. For each dataset, we first individually evaluate the emotion-classification performance obtained by each modality. We then evaluate the performance obtained by fusing the features from these modalities. We show that our algorithms outperform the results reported by other studies for emotion/valence/arousal/liking classification on DEAP and MAHNOB-HCI datasets and set up benchmarks for the newer AMIGOS and DREAMER datasets. We also evaluate the performance of our algorithms by combining the datasets and by using transfer learning to show that the proposed method overcomes the inconsistencies between the datasets. Hence, we do a thorough analysis of multi-modal affective data from more than 120 subjects and 2,800 trials. Finally, utilizing a convolution-deconvolution network, we propose a new technique towards identifying salient brain regions corresponding to various affective states.
\end{abstract}


\IEEEpeerreviewmaketitle

\section{Introduction}
\IEEEPARstart{I}{n} recent years, there has been growing interest towards approaching research in affective computing from bio-sensing perspective. To be sure, it is not just in affective computing that research in bio-sensing has been gaining popularity. Other avenues of research such as health \cite{health,health_2}, virtual reality \cite{VR}, robotics \cite{EEG_Quadcopter,Robotics}, content rating \cite{content_rating}, etc. have also exploited bio-sensing as a research tool. Bio-sensing systems specifically those which are used to measure electrocardiogram (ECG), electroencephalogram (EEG), galvanic skin response (GSR), etc. have been around for decades. But, because of their bulkiness and complexity, they were restricted to controlled laboratory environments and hospitals. The current interest in utilizing bio-sensing systems for various applications has been motivated or driven by the development of wearable bio-sensing systems that make data collection faster and easier \cite{emotiv,earPPG_MIT,ecg_devices}. The advances in hardware have led to the further development of multi-modal bio-sensing systems i.e. those capable of monitoring and recording multiple bio-signals simultaneously \cite{Biovotion,imotions,my_hcii,my_tbme}.

Many research studies have shown that it is possible to recognize human emotions by the use of facial expressions from images and videos \cite{face_AUs,face_only,face_only_2,multi-modal-without-biosensing}. Advances in deep-learning have also made it possible to train large neural networks on big datasets for research in affective computing \cite{video_DL_emotion,video_DL_emotion_2,video_DL_emotion_3} apart from other problems such as object detection and classification \cite{YOLO,CIFAR,MNIST}. Compared to the amount of deep-learning research that has translated towards solving problems involving images/videos, the deep-learning research conducted on bio-sensing data has been sparse. A recent survey on using EEG for affective computing \cite{affective_computing_survey} suggests that in almost all the cases the feature extraction and classification steps do not utilize deep neural networks.

\begin{table*}[!ht]
\centering
\caption{Table highlighting the inconsistencies among the datasets and sensing modalities used in this study}
\label{table-datasets}
\resizebox{\textwidth}{!}{%
\begin{tabular}{p{1.75in}p{1.75in}p{1.75in}p{1.75in}}
\\\hline\textbf{DEAP Dataset \cite{DEAP}} & \textbf{AMIGOS Dataset \cite{AMIGOS}} & \textbf{MAHNOB-HCI Dataset \cite{MAHNOB-HCI}} & \textbf{DREAMER Dataset \cite{Dreamer_Database}} \\\hline                           

\\32 subjects & 40 subjects & 27 subjects & 23 subjects                                                                                 
\\[0.2cm]40 trials using music videos (trial length fixed at 60 seconds) & 16 trials using movie clips (trial length varying between 51 and 150 seconds) & 20 trials using movie clips (trial length varying between 34.9 and 117 seconds) & 18 trials using movie clips (trial length varying between 67 and 394 seconds)     
\\[0.5cm]Raw and pre-processed data available & Raw and pre-processed data available & Only raw data available & Only raw data available
\\[0.5cm]32-channel EEG system (Two different EEG systems used. Channel locations: Fp1, AF3, F7, F3, FC1, FC5, T7, C3, CP1, CP5, P7, P3, Pz, PO3, O1, Oz, O2, PO4, P4, P8, CP6, CP2, C4, T8, FC6, FC2, F4, F8, AF4, Fp2, Fz, Cz) & 14-channel EEG system (A single EEG system used for all subjects. Channel locations: AF3, F7, F3, FC5, T7, P7, O1, O2, P8, T8, FC6, F4, F8, AF4) & 32-channel EEG system (A single EEG system used for all subjects. Channel locations: Fp1, AF3, F7, F3, FC1, FC5, T7, C3, CP1, CP5, P7, P3, Pz, PO3, O1, Oz, O2, PO4, P4, P8, CP6, CP2, C4, T8, FC6, FC2, F4, F8, AF4, Fp2, Fz, Cz) & 14-channel EEG system (A single EEG system used for all subjects. Channel locations: AF3, F7, F3, FC5, T7, P7, O1, O2, P8, T8, FC6, F4, F8, AF4)
\\[0.2cm]--- & 2-channel ECG system & 3-channel ECG system & 2-channel ECG system                                         
\\[0.2cm]1-channel PPG system & --- & --- & ---                                                 
\\[0.2cm]1-channel GSR system & 1-channel GSR system & 1-channel GSR system & ---    

\\[0.2cm]Sampling rate 128 Hz & Sampling rate 128 Hz & Sampling rate 256 Hz & Sampling rate EEG/ECG: 128/256 Hz

\\[0.2cm]Face video recorded for 22 of 32 subjects (EEG cap and EOG electrodes occludes parts of the forehead and cheeks) & Face video recorded for all subjects (Only a small portion of the forehead is occluded by the EEG system) & Face video recorded for all subjects (Only a small portion of the forehead is occluded by the EEG system) & --- 
\\[1.2cm]3-seconds of pre-trial baseline data available. & No baseline data available. & 30 seconds of pre-trial and post-trial baseline data available. & 61 seconds of pre-trial baseline data available
\\[0.2cm]Valence/Arousal/Liking rated using a continuous scale between 1 to 9 & Valence/Arousal/Liking rated using a continuous scale between 1 to 9 & Valence/Arousal rated using a discrete scale of integers from 1 to 9 & Valence/Arousal rated using a discrete scale of integers from 1 to 5 \\\hline
\end{tabular}%
}
\end{table*}

There are chiefly three reasons limiting the use of deep-learning to bio-sensing modalities. First, it is easier to create an image/video database by collecting a huge amount of image/video data with any decent camera (even that of a smartphone) whereas the data collection of bio-signals is often costly, time-consuming, and laborious. Second, the image/video datasets generated using different cameras are usually consistent or can easily be made so by changing the frame resolution or modifying the number of frames being captured per second without losing critical information in the process. On the other hand, commercially available bio-sensing devices vary widely in terms of sampling rate, analog to digital resolution, numbers of channels and sensor positioning \cite{ecg_devices,affective_computing_survey}. Furthermore, there are differences in the signal profiles between different types of bio-sensing signals such as EEG vs ECG. Third, visualizing image data for object detection or assessing emotions by looking at faces/body postures in the images is much easier (such as for manual tagging) and intuitive. But, extracting meaningful knowledge about various features from bio-sensing signals requires pre-processing. Unlike image data, additional steps are required in bio-sensing data to first filter the data of any noise such as due to motion artifacts or unwanted muscle activity.

Using multiple bio-sensing modalities can be advantageous over using a singular one because the salient information in the respective modalities may be independent of and complementary to each other to some extent. Thus, together they may enhance the performance for a given classification task \cite{GSR_and_PPG,MPantic_multimodal_emotions}. In most cases, the emotion-classification problem has been approached by measuring the arousal and valence as given by the emotion circumplex model \cite{russell}. It is evident from various studies \cite{affective_computing_survey,deap_22subj,deap_korea_1} that a single modality may perform differently for arousal and valence classification. So, in theory, two modalities that show good performance independently for valence and arousal respectively, may perform even better jointly for the emotion classification problem.

This study focuses on multi-modal data from both bio-sensing and vision-based perspectives. We fuse the features with deep-learning-based methods and traditional algorithms for all modalities on four different datasets. We show that using multi-modality is advantageous over singular modalities in various cases. Finally, we show that the deep-learning methods perform well even when the size of the dataset is small. For each of the four datasets, we show that our methods outperform previously reported results. To the best of our knowledge, this study contains the most exhaustive use of multi-modal bio-sensing data for affective computing research. Our results also demonstrate the applicability of deep-learning-based methods to overcome the discrepancies between different modalities and even effectively fuse the information from them, as shown by results from combining the datasets and transfer learning.

\begin{table*}[!ht]
\centering
\caption{Classification performance and evaluation performed by various reported studies on the four datasets}
\label{table-DEAP}
\resizebox{\textwidth}{!}{%
\begin{tabular}{p{1in}p{1in}p{1in}p{1in}p{3in}}
\\\hline
\textbf{Study} & \textbf{Used Modalities}            & \textbf{Extracted Features}                                         & \textbf{Classifier}             & \textbf{Evaluation} \\\hline

\\[0.05cm]
\multicolumn{5}{c}{\textbf{DEAP Dataset}}
\\[0.2cm]
\textbf{Liu et al. \cite{deap_22subj}}    & EEG                          & Fractal dimension (FD) based                              & SVM                             & Only 22 of the 32 subjects used. 50.8\% Valence (4-classes) and 76.51\% Arousal/Dominance.                                                     

\\[0.2cm]
\textbf{Yin et al. \cite{deap_yin}}    & EEG, ECG, EOG, GSR, EMG, Skin temperature, Blood volume, Respiration        & Various                                       & MESAE                           & 77.19\% Arousal and 76.17\% Valence (2-classes) using fusion of all modalities.  

\\[0.2cm]
\textbf{Patras et al. \cite{DEAP}} & EEG                          & PSD                                                       & Bayesian Classifier             & 62\% Valence and 57.6\% Arousal (2-classes)                                                                                                                                
\\[0.2cm]
\textbf{Chung et al. \cite{chung_DEAP}}  & EEG                          & Various & Bayesian weighted-log-posterior & 70.9\% Valence and 70.1\% Arousal (2-classes)                                                                                                                                                        
\\[0.6cm]
\textbf{Shang et al. \cite{Shang_DL_DEAP}}  & EEG, EOG, EMG                          &   Raw data                                     & Deep Belief Network, Bayesian Classifier             & 51.2\% Valence, 60.9\% Arousal, and 68.4\% Liking (2-classes)                                                                                                                                    
\\[0.6cm]
\textbf{Campos et al. \cite{Campos_DEAP}} & EEG                          & Various                                                   & Genetic algorithms, SVM         & 73.14\% Valence and 73.06\% Arousal (2-classes)

\\\hline

\\[0.05cm]
\multicolumn{5}{c}{\textbf{AMIGOS Dataset}}
\\[0.2cm]
\textbf{Miranda et al. \cite{AMIGOS}} & EEG, ECG, GSR  & Various                                                   & SVM         & \textsuperscript{\textasteriskcentered}57.6/53.1/53.5/57 Valence and 59.2/54.8/55/58.5 Arousal (2-classes) using EEG/GSR/ECG alone/EEG, GSR, and ECG fusion.
\\\hline  

\\[0.05cm]
\multicolumn{5}{c}{\textbf{MAHNOB-HCI Dataset}}
\\[0.2cm]
\textbf{Soleymani et al. \cite{MAHNOB-HCI}} & EEG, ECG, GSR, Respiration, Skin Temperature  & Various                                                   & SVM         & 57/45.5/68.8/76.1\% Valence and 52.4/46.2/63.5/67.7\% Arousal (2-classes) using EEG/Peripheral/Eye gaze/Fusion of EEG and gaze.

\\[0.2cm]
\textbf{Koelstra et al. \cite{Koelstra_Mahnob}} & EEG, Faces  & Various                                                   & Decision classifiers fusion         & 73\% Valence and 68.5\% Arousal (2-classes) using EEG and Faces fusion.

\\[0.2cm]
\textbf{Alasaarela et al. \cite{Alasaarela_Mahnob}} & ECG  & Various & KNN         & 59.2\% Valence and 58.7\% Arousal (2-classes)

\\[0.2cm]
\textbf{Zhu et al. \cite{Zhu_Mahnob}} & EEG and Video stimulus & Various & SVM         & 55.72/58.16\% Valence and 60.23/61.35\% Arousal (2-classes) for EEG alone/Video stimulus as privileged information with EEG.

\\\hline  

\\[0.05cm]
\multicolumn{5}{c}{\textbf{DREAMER Dataset}}
\\[0.2cm]
\textbf{Stamos et al. \cite{Dreamer_Database}} & EEG, ECG  & PSD, HRV                                                   & SVM         & 62.49/61.84\% Valence and 62.17/62.32\% Arousal (2-classes) using EEG alone/EEG and ECG fusion.
\\[-0.3cm]
\\\hline                                    
\end{tabular}%
}
\normaljustify
\vspace{1ex}

\textsuperscript{\textasteriskcentered}Denotes mean F1-score. Accuracy value not available.
\end{table*}

\section{Materials and Related Work}
We designed and evaluated our framework on four publicly available multi-modal bio-sensing and vision-based datasets namely DEAP \cite{DEAP}, AMIGOS \cite{AMIGOS}, MAHNOB-HCI \cite{MAHNOB-HCI}, and DREAMER \cite{Dreamer_Database}. Table \ref{table-datasets} briefly describes the four datasets, with a focus on the modalities that we used in this work. For DEAP and AMIGOS datasets, we used the preprocessed bio-sensing data that has been suitably re-sampled and filtered whereas for MAHNOB-HCI and DREAMER datasets we perform the filtering and artifact removal before extracting features. The trials in the DEAP and AMIGOS datasets have been tagged by subjects for valence, arousal, liking, and dominance on a continuous scale of 1 to 9. For MAHNOB-HCI and DREAMER datasets, the valence and arousal have been tagged on a discrete scale using integers from 1 to 9 and 1 to 5, respectively. We used the emotion circumplex model \cite{russell} to divide the emotions into four categories namely, High-Valence High-Arousal (HVHA), Low-Valence High-Arousal (LVHA), Low-Valence Low-Arousal (LVLA), and High-Valence Low-Arousal (HVLA). These categories loosely map to happy/excited, annoying/angry, sad/bored, and calm/peaceful emotions, respectively. For each dataset, the labels were self-reported by the subjects after the presentation of the video stimuli.

As shown in Table \ref{table-datasets}, the datasets differ in many aspects. Hence, many traditional algorithms cannot be generalized across datasets because of differences in the number and nature of extracted features. Apart from the types of the audio-visual stimulus (music videos vs. movie clips), the datasets vary in the trial duration and baseline data availability. The DEAP dataset has trial length fixed at 60 seconds whereas for the AMIGOS dataset, the trial length varies between 51 to 150 seconds. This varying trial length is significant since the longest video is about thrice the length of the shortest one. Hence, if a particular emotion of the subject is invoked for 25 seconds during a trial, it will appear in half of the trial in the shortest video but only in one-sixth of the trial in the longest one. Furthermore, the trial length variation of the DREAMER dataset is even greater than that in the AMIGOS dataset. There is also no baseline data present in the AMIGOS dataset to compensate for the subject's initial emotional power (defined as the distance from the origin in the emotion circumplex model). Different kinds of systems have been used to collect the EEG data in these datasets. 32-channel EEG in the DEAP and MAHNOB-HCI datasets may contain much more emotion-relevant information than the 14-channel EEG in the AMIGOS and DREAMER datasets.

Only the DEAP dataset uses photoplethysmogram (PPG) to measure heart rate instead of ECG. The use of PPG generally loses the information that is present in the ECG waveform such as QRS complex, PR, and ST segment lengths, etc.

The EEG electrodes introduce varying degrees of occlusion while capturing frontal videos of the subjects. This effect was found to be more problematic in the DEAP dataset because of the placements of the EOG electrodes on subjects' faces. Furthermore, some data are missing in some modalities in a subset of the trials of these datasets. We ignore such missing entries in our evaluation.

Table \ref{table-DEAP} shows that in almost all the cases EEG has been the preferred bio-sensing modality while vision modality i.e. the use of the frontal videos to analyze facial expressions has not been commonly used on these datasets. The classification accuracy for all emotion classes as per the circumplex model rather than only for arousal/valence are rarely reported. In other cases such as \cite{deap_korea_1,deap_korea_2}, where the analysis of emotions is reported, the goal seems to be clustering the complete dataset into four classes rather than having a distinct training and testing partition for evaluation.

In terms of accuracy, we see from Table \ref{table-DEAP} that using multiple sensor modalities, the best performance on the DEAP datset is by \cite{deap_yin} when utilizing data from mulitple modalities. For the MAHNOB-HCI dataset, the best accuracy for valence and arousal is 73\% and 68.5\% respectively \cite{Koelstra_Mahnob}, which is again using multiple sensor modalities. The AMIGOS and DREAMER datasets were released recently and hence only baseline evaluation on these have been reported in Table \ref{table-DEAP}.

This study will utilize complete datasets and not a subset of them, as in some previous studies. We evaluate our methods with disjoint partitions between training, validation, and test subsets of the complete datasets. Our evaluation is first reported for all modalities separately (including using frontal videos that were ignored by other studies) and then combining them together. Since not all datasets and previous studies report results on Dominance, we chose to classify valence, arousal, liking and emotions as the affective measures.

\section{Research Methods}
In this section, we detail various types of methods that we employed to extract features from each bio-sensing modality and frontal videos. 

\subsection{EEG feature extraction}
For the DEAP and AMIGOS datasets, preprocessed EEG data are available, bandpass-filtered between 4-45 Hz, and corrected for eye-blink artifacts. For the MAHNOB-HCI and DREAMER datasets, we performed the bandpass filtering and artifact removal using the Artifact Subspace Reconstruction (ASR) toolbox \cite{ASR} in EEGLAB \cite{EEGLAB}. The processed EEG data were then converted into the frequency domain to extract both traditional and deep-learning based features (see below).

\subsubsection{EEG-PSD features}
For each EEG channel, we extracted the traditional power spectral density (PSD) in three EEG bands namely, theta (4-7 Hz), alpha (7-13 Hz), and beta (13-30 Hz). These EEG bands were chosen since they account most towards human cognition. We used half second overlapping windows. The PSD was then averaged over the total trial length. Hence, because of the differences in the number of EEG channels, we get 96 features for trials in the DEAP and MAHNOB-HCI datasets and 42 features for trials in the AMIGOS and DREAMER datasets.

\subsubsection{Conditional entropy features}
To get information regarding the interplay between different brain regions, we extract conditional entropy-based features. The conditional entropy between two random variables carries information about the uncertainty in one variable given the other. Hence, it acts as a measure of the amount of mutual information between the two random variables. The mutual information $I(X;Y)$ of discrete random variables $X$ and $Y$ is defined as 

\begin{equation}
I(X;Y) = \sum_{x\in X}\sum_{y\in Y}p(x,y) log\bigg(\frac{p(x,y)}{p(x)p(y)}\bigg)
\end{equation}

The conditional entropy will be zero if the signal Y is completely determined by signal X. To calculate the conditional entropy, we first calculated the mutual information $I(X;Y)$ between the two signals, which requires the calculation of the approximate density function $\hat p(x)$ of the following form

\begin{equation}
\hat p(x) = \frac{1}{N} \sum_{i=1}^{N} \delta(x - x^{(i)}, h)
\end{equation}

where $\delta(.)$ is the Parzen window, $h$ is the window width, $N$ is the samples of variable $x$ and $x^{(i)}$ is the $i$th sample. This approximate density function is calculated as an intermediary step to the calculation of true density $p(x)$, since when N goes to infinity it can converge to the true density if $\delta(.)$ and $h$ are properly chosen \cite{conditional_entropy}. We chose $\delta(.)$ as the Gaussian window

\begin{equation}
\delta(z,h) = exp\bigg(-\frac{z^T \Sigma^{-1} z}{2 h^2}\bigg) \bigg/ \Big\{ (2\pi)^{d/2} h^d |\Sigma|^{1/2} \Big\}
\end{equation}

where $z = x - x^{(i)}$, $\Sigma$ is the covariance of $z$, and $d$ is the dimension of $x$. By plugging the value of $d = 1, 2$ we get the marginal density $p(x)$ and the density of the bivariate variable $(x,y),p(x,y)$ \cite{conditional_entropy}. In this manner the mutual information $I(X;Y)$ is calculated which is related to conditional entropy $H(Y|X)$ by

\begin{equation}
I(X;Y) = H(Y) - H(Y|X)
\end{equation}

The conditional entropy between all possible pairs of EEG channels was calculated over the complete trial length \cite{my_EMBC}. Hence, due to the differences in the number of EEG channels, 496 conditional entropy features were calculated for the DEAP and MAHNOB-HCI datasets and 91 features for the AMIGOS and DREAMER datasets.

\subsubsection{EEG-PSD images-based Deep Learning features}
\begin{figure}[!ht] \centering
{\includegraphics[width=3.45in]{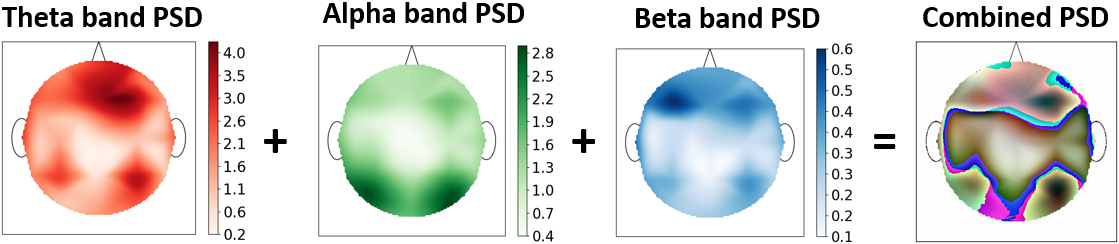}}
\vspace{-15pt}
\caption{PSD heat-maps of theta (\textcolor{red}{red}), alpha (\textcolor{green}{green}), and beta (\textcolor{blue}{blue}) EEG bands being added according to respective color-bar range to get combined RGB heat-map (Image border, nose, ears, and color-bars have been added for visualization only.)}
\label{fig:bands-added-image}
\end{figure}

In this section, we propose a novel method for feature extraction from the EEG data, which is based on deep convolution networks without requiring a large amount of training data. The method can also work in a similar manner for different types of EEG datasets i.e. datasets with different numbers and placements of electrodes, sampling rates, etc. We first used the computed EEG-PSD features from the first method mentioned above to plot power spectrum heat maps for the three EEG bands using bicubic interpolation to calculate the values in the 2-D plane. These images now contain the topographical information for the three frequency bands according to the standard EEG 10-20 system (Fig. \ref{fig:bands-added-image}). It is worth noting that the commonly used EEG-PSD features in themselves do not take into account the EEG-topography i.e. the locations of EEG electrodes for a particular EEG band. Hence, we try to exploit EEG-topography to extract information regarding the interplay between different brain regions. It is for exploiting this information that we convert EEG data to an image-based representation and utilize pre-trained deep-learning networks to extract such relationship between various brain regions.

As shown in Fig. \ref{fig:bands-added-image}, we used `Red' color-map for the theta band, `Green' color-map for the alpha band, and `Blue' color-map for the beta band. We then combined the three colored images into an RGB colored image \cite{RGB_DL}. Based on the ranges and maximum values in the data for the three EEG bands, we used the ratio of alpha blending \cite{MatPlotLib} to give weights to the three individual bands' images before adding them together. This color image carries information about how the power in the three bands interacts with each other across the different brain regions. For example, a yellow colored region has a higher amount of power in the theta (red) and alpha (blue) bands, whereas a pink colored region has high power in the theta (red) and beta (blue) bands. In this manner, a single brain heat-map image can be used to represent spatial and spectral information in the three EEG bands. That is, we obtained one image representing the topographic PSD information for every trial. Because the images in the four datasets can be formed in a similar manner irrespective of the different numbers of EEG channels and positions, we can use this method across the four datasets easily.

We then used a pre-trained VGG-16 network \cite{VGG} to extract features from the combined RGB heat-map image. This network consists has been trained with more than a million images for 1,000 object categories using the Imagenet Database \cite{ImageNet}. It has been shown that a pre-trained VGG-16 network can be utilized for feature extract and classification for applications different than it was trained for \cite{VGG_2,my_itsc}. The RGB image is resized to $224{\times}224{\times}3$ size before submitting to the network. The last but one layer of this VGG network consists of 4,096 most significant features, which we extracted for emotion classification. Principal component analysis (PCA) was then applied to this feature space to reduce its dimension to 30 for each trial \cite{PCA}. We then combined the features from this method and from the conditional entropy mentioned above for evaluation.

The resultant EEG-PSD images from the above method can be used to denote how the EEG spectral activity is distributed across various brain regions across time for a particular kind of stimulus. This can be done by sending such successive images (varying across time) to a reverse deep-learning network and detect the most salient features i.e. activated regions of the brain across time for a particular stimulus. In the evaluation section below, we utilize these brain images to denote the brain regions that are most activated for different affective responses.

\subsection{ECG/PPG-based feature extraction}
Both ECG and PPG signals can be used to measure heart rate (HR) and heart-rate variability (HRV), though ECG can provide more useful information due to its greater ability to capture ECG t-wave etc. For consistency between the two types of signal measurements i.e. PPG and ECG, we employ two methods in the same manner on data from both of these modalities in the four datasets.

\subsubsection{HRV features}
\begin{figure}[!ht] \centering
{\includegraphics[width=3.45in]{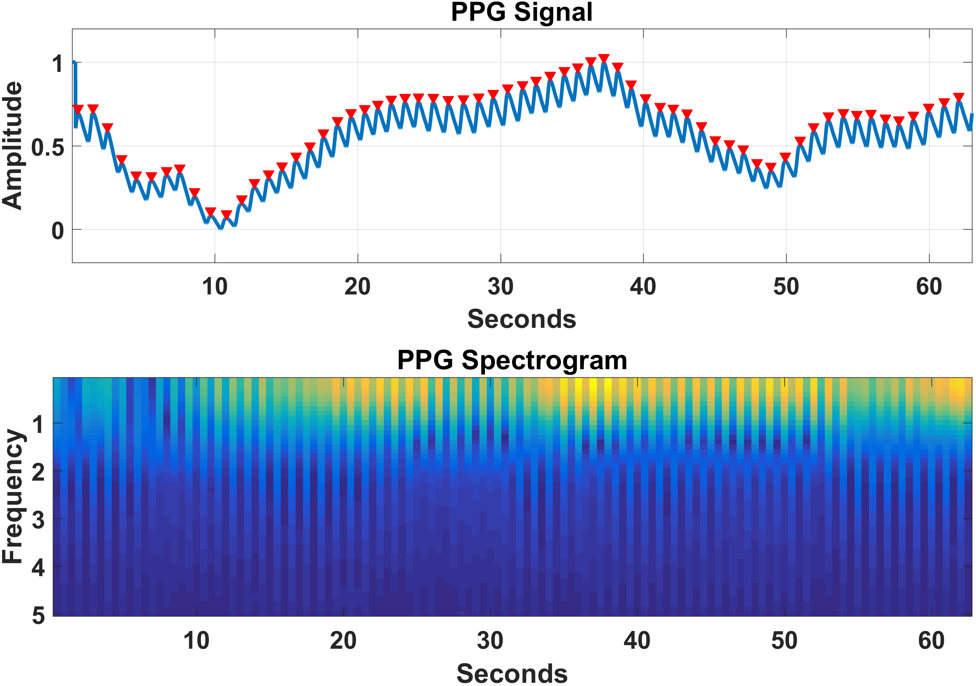}}
\caption{For a trial from DEAP dataset, PPG signal with peaks (in \textcolor{red}{red}) being detected for the calculation of RRs and HRV (above), and PPG spectrogram (below).}
\label{fig:ppg-plot}
\end{figure}

HRV has shown to be a good metric for classifying emotional valence and arousal \cite{hrv_only}. For every trial (whether PPG in the DEAP dataset or ECG in the AMIGOS, MAHNOB-HCI, and DREAMER datasets), we first used a moving-average filter with a window length of 0.25 seconds to filter out the noise in the data. We then used a peak-detection algorithm \cite{peak_detect} after scaling the data between 0 and 1. The minimum distance between successive peaks as being at least 0.5 seconds apart was taken as the threshold to remove false positives as in Fig. \ref{fig:ppg-plot}.

\begin{figure*}[!ht] \centering
{\includegraphics[width=\textwidth, height = 1.7in]{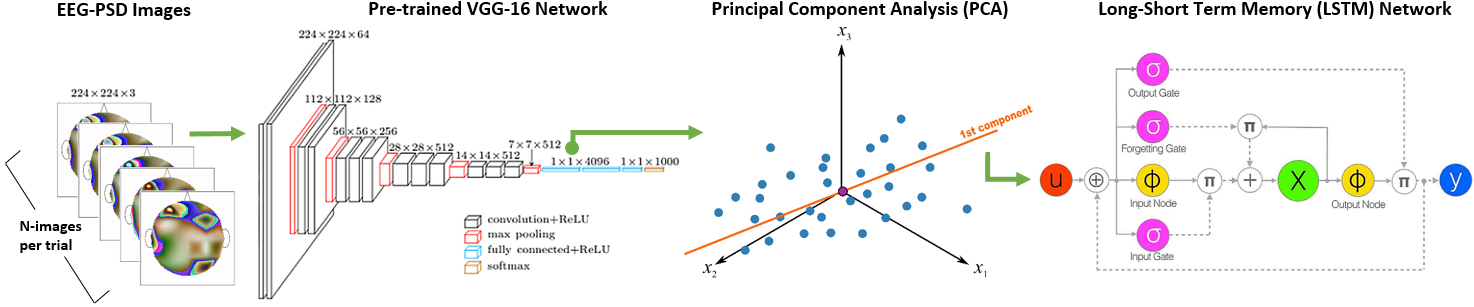}}
\vspace{-15pt}
\caption{Network architecture for EEG-PSD trend based deep-learning method.}
\label{fig:eeg_method_6}
\end{figure*}

The total number of peaks per minute represents the subject's heart rate. To calculate the HRV, the time differences between successive peaks were calculated to get inter-beat intervals (RRs). These RRs were then used to compute HRV using the pNN50 method \cite{pNN50_algorithm}. This method of HRV calculation measures the percentage of successive RR intervals that differ by more than 50ms. This method has been shown to be correlated with the activity of the parasympathetic nervous system (PNS) \cite{PNS_Activity_Paper}. For the datasets containing multiple ECG channels, we performed the same procedure for all the channels.

\subsubsection{Extracting deep-learning-based spectrogram features of ECG/PPG}
Previous studies have reported that frequency-domain features in the ECG work well for tasks such as ECG beat discrimination \cite{ECG_Frequency}. To exploit time-frequency information from ECG/PPG, we extracted deep-learning based features on ECG/PPG by converting the time-series data to a frequency domain-based image representation. The frequency range of ECG/PPG signals is low and hence we focus only on 0-5 Hz range. We generated a spectrogram \cite{spectrogram} over the complete trial in this frequency range as in Fig. \ref{fig:ppg-plot}. To get a good amount of variations, we chose Parula color map for the spectrogram image. The frequency bins with various colors at different frequencies represent the signal across the trial length. We employed the same procedure to get the spectrogram images of ECG/PPG signals from the four datasets. We then resized the spectrogram images to feed them into the VGG-16 network, and after which the resultant 4,096 extracted features were reduced to 30 features using PCA. The features from this method were concatenated with the HRV features from above for evaluation.

\subsection{GSR-based feature extraction}
Similar to ECG/PPG, we employ two methods to extract features from the GSR data, one in the time domain and the other in the frequency domain.

\subsubsection{Statistical features}
We first used a moving average filter with a window length of 0.25 seconds to remove noise from the GSR signal. For each trial, we calculated eight statistical features from the time-domain GSR data. Two features are the number of peaks and the mean of absolute heights of the peaks in the signal. Six more statistical features based on $n^{th}$ order moments of the GSR time-series data were calculated as shown in \cite{gsr_features}. These features measure the trend i.e. variations in the GSR data in actual, and successive first and second differences of the signal.


\subsubsection{Extracting deep-learning-based spectrogram features of GSR}
GSR signals change very slowly and hence we focus only on the 0-2 Hz frequency range. Similar to ECG, we generated the spectrogram image of GSR for each trial in the above frequency range. We then extracted VGG-16 network-based features that characterize the most meaningful interactions between various edges in the spectrogram. These features were reduced to 30 using PCA and then concatenated with the time-domain GSR features from above.

\subsection{Frontal video-based feature extraction}
Unlike other studies in Table \ref{table-DEAP}, we also use the frontal videos of the subjects for emotion/valence/arousal/liking classification. Facial expressions can be very reliable indicators of one's emotions based on his/her personality i.e. willingness to show emotions by various facial expressions. For each frontal video trial, we first extracted a single frame for every second in the trial by extracting the first frame for every second of the video. We excluded the extreme ends of the image and placed a threshold on the minimum face size to be $50{\times}50$ pixels. This was done to reduce computational complexity and increase face detector's accuracy. Face detection was done using Haar-like features based on Viola-Jones object detector \cite{Viola_Jones}. A small portion of images had a majority of the face occluded due to subject putting his/her hand over their face. The face detector failed on these instances and hence these were discarded.

\begin{figure}[!ht] \centering
{\includegraphics[width=3.45in]{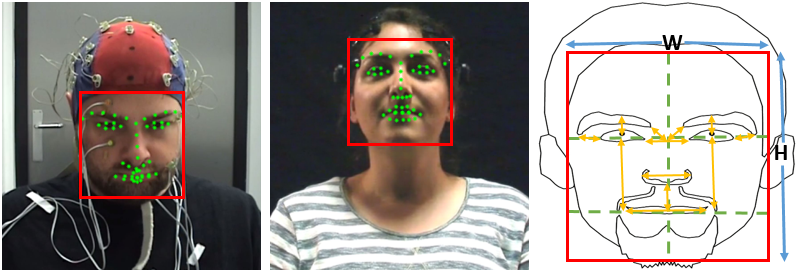}}
\vspace{-15pt}
\caption{Detected face (marked in \textcolor{red}{red}) and face localized points (marked in \textcolor{green}{green}) in DEAP Dataset (left), AMIGOS Dataset (center), and subset of features (marked in \textcolor{yellow}{yellow}) computed using face localized points (right). The features are normalized using height (H) and width (W) of the detected face. These subjects' consent to use their face is marked in respective datasets.}
\label{fig:face-features}
\end{figure}

\subsubsection{Facial points localization based features}
We applied the state-of-the-art Chehra algorithm \cite{chehra} to the extracted facial regions to obtain 49 localized points on the face representing the significant parts as shown in Fig. \ref{fig:face-features}. This algorithm does not need any human input or a dataset-specific training model for predicting the localized face points, making the process fully automated. Previous research studies have reported promising results using the face action units (AUs) based on such facial landmarks \cite{face_AUs}. We used these 49 localized points to calculate 30 features based on distances such as that between the center of the eyebrow to the center of the eye, between the nose and the middle part of the upper lip, between upper and lower lips, etc. Many of the 30 features are described in \cite{face_AUs} while others by designed by hand. All such features were normalized based on the height and width of the detected face to remove variations due to the distance from the camera. The mean, $95^{th}$ percentile (more robust than maximum), and standard deviation of these 30 features across the frames in a single trial are then calculated. These 90 parameters were then used for evaluation.

\subsubsection{Deep Learning-based features}
The use of deep-learning has transformed computer vision in multiple ways. This is because such deep networks are capable of extracting feature representations from images that capture both uniform (contrast etc.) and complex (small changes in texture etc.) types of features. Hence, we utilized these networks on face-images using a deep network pre-trained on VGG-faces dataset \cite{VGG_Faces}. The extracted face region was resized to $224{\times}224{\times}3$ for each selected frame in the trial. Similar to the CNN-based deep-learning method used above for the bio-sensing modalities, we extracted 4,096 most meaningful features on these resized images. But, unlike the bio-sensing method, we employed a different VGG network that has been specifically trained on more than 2.6 million face images from more than 2,600 people \cite{VGG_Faces}. This was done to extract features that are more relevant to the face-dependent feature space. The mean, $95^{th}$ percentile, and standard deviation of the features across the images in every trial were computed and the subsequent features space was reduced to 30 using PCA.

\subsection{Dynamics of the EEG/Face features using deep-learning}
The above-mentioned methods for extracting deep-learning features from EEG/face-videos are special cases in which a single trial is represented by a single image (EEG-PSD image/Single feature space for face images in a video). But, these methods do not fully take into account the temporal dynamics of the features over time within the trial. Hence, we propose a new method in which such images (EEG-PSD or face region) are utilized for every second within a trial. Fig. \ref{fig:eeg_method_6} shows the network architecture for this method for the EEG-PSD images. Multiple EEG-PSD images were formed for each trial by generating one image for each second, all of which went through the pre-trained VGG network. The 4,096 features from the off-the-shelf deep-learning network were then obtained for each image. In addition, the conditional entropy features for every second were also calculated. PCA was then used to reduce the dimensionality of the feature space comprising features from EEG and face-videos. The resultant feature space has 60 most representative features. These $60{\times}N$ ($N =$ trial length in seconds) features are then sent to a Long-Short Term Memory (LSTM) network \cite{LSTM}. However, this method could only be employed on the DEAP dataset since the AMIGOS and MAHNOB-HCI datasets have varying trial length and DREAMER dataset does not contain any video data. The huge variations in the trial length in the AMIGOS and MAHNOB-HCI datasets meant that during the data preparation phase of LSTM, a large amount of padding was needed. This may be possible in data from physical sensors (like temperature, luminous, pressure, etc.) where interpolation is easy to perform. But, for bio-signals, this is not desirable because we do not have affective labels reported by the subject during the course of each video trial i.e. we do not know which parts of the video contributed most towards the affective response. Hence, we could not use LSTMs on these datasets.

\section{Evaluation}
In this section, we evaluate the various feature-extraction methods described above. First, we compare the performance of the classification of affective states using the deep-learning features from the pre-trained convolution network with that using traditional EEG features. We also report the classification performance when features from these modalities are fused together. Thereafter, we evaluate the classification performance using each modality individually on the four datasets, on combining the datasets together, and for transfer learning. Finally, we present results for a novel deep-learning based technique to identify the most important brain regions associated with emotional responses.

\begin{figure}[!ht] \centering
{\includegraphics[width=3.45in]{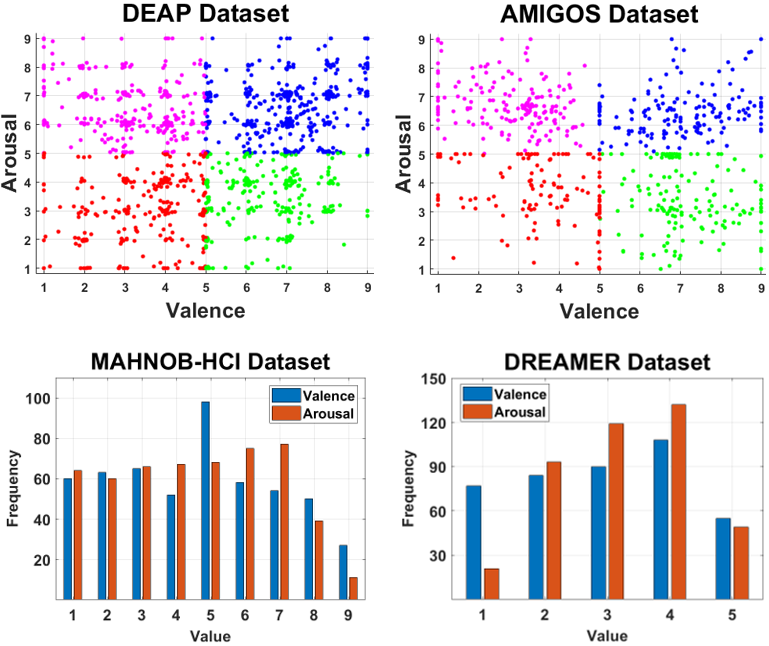}}
\vspace{-15pt}
\caption{Distribution of emotion classes in the four datasets.}
\label{fig:va-datasets}
\end{figure}

Fig. \ref{fig:va-datasets} shows the distribution of self-reported valence and arousal for the four datasets. It is evident that the DEAP dataset has a higher concentration of trials closer to neutral emotion i.e. near the center of the graph. For each individual dataset separately, we perform leave-one-subject-out evaluation and show results for single modality classification in Table \ref{table-singular-modality} and multi-modality classification in Table \ref{table-multi-modality}. Then, we performed an evaluation by combining datasets together and show results in Table \ref{table-combined-dataset}. Finally, we used transfer learning among the datasets i.e. training on one dataset and testing on another (Table \ref{table-transfer-dataset}). For these two latter evaluations of combining the datasets and using transfer learning, we randomly divide the datasets into two parts with an 80/20 ratio and perform 10-fold cross-validation. The classification was done using extreme learning machines (ELM) \cite{ELM} with variable numbers of neurons, which has been shown to perform better than support vector machines (SVM) for various cases \cite{deap_hong_kong_1}. All the features were re-scaled between -1 and 1 before training the ELM. A single-layer ELM was used with a sigmoid activation function. For the trend-based deep-learning method, two hidden layers in the LSTM were used with the number of neurons being 200 and 100, respectively. Stochastic gradient descent with a momentum (SGDM) optimizer was used to train the LSTM network.

\subsection{Visualizing class-separability using the traditional vs. deep-learning features}
\begin{figure}[!ht] 
\centering
{
\subfloat[t-SNE on two valence classes (low-valence in \textcolor{blue}{blue} and high-valence in \textcolor{red}{red}) ]{
\includegraphics[width=3.45in]{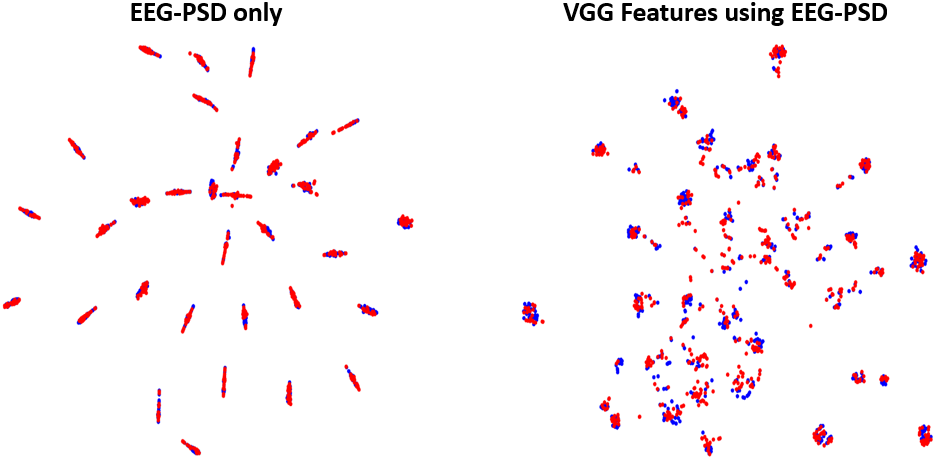}}

\subfloat[t-SNE on four emotion classes (HVHA in \textcolor{blue}{blue}, LVHA in \textcolor{red}{red}, LVLA in \textcolor{magenta}{magenta}, and HVLA in \textcolor{green}{green})]{
\includegraphics[width=3.45in]{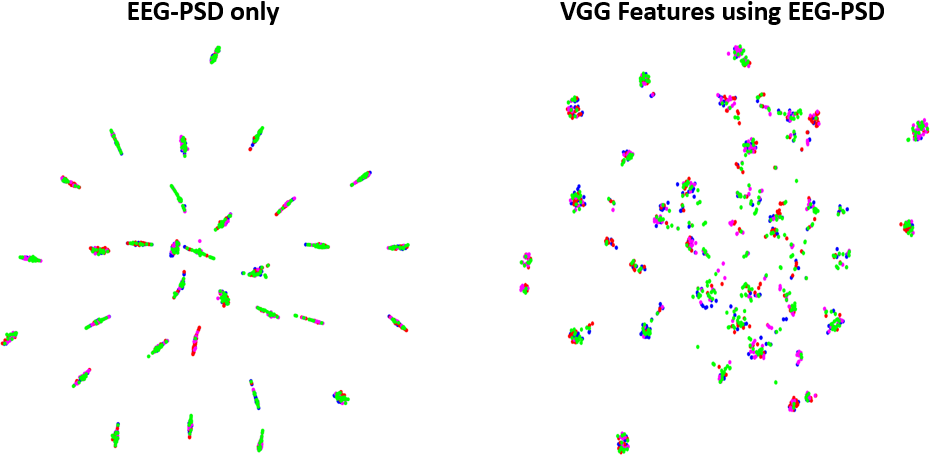}}
}
\caption{Visualization of feature spaces using t-SNE \cite{t-sne} in trials from the DEAP dataset on the EEG-PSD features and the VGG features derived from the combined RGB image. The VGG features allow for better separation.}
\label{fig:features-compare}
\end{figure}

One of the hypotheses of our study is that the traditional methods for analyzing EEG can be improved by using deep-learning based features obtained from pre-trained convolution networks. This is important because training convolution networks requires huge datasets, which are usually unavailable in the bio-sensing domain. Hence, the Deep-Learning method described in section \rmnum{3}.A3 should be able to extract more meaningful features from EEG-PSD features (\rmnum{3}.A1). We used t-SNE \cite{t-sne} to visualize the dimensionally reduced space using traditional EEG-PSD features for 2-class valence and 4-class emotions on the DEAP dataset with fixed trial length. Kullback-Leibler (KL) divergence was used for measuring similarity and Euclidean distance was used as the distance measure for the t-SNE implementation. We then applied the same approach to the features obtained by the VGG network, which were computed after using the EEG-PSD features to create a combined RGB image in Fig. \ref{fig:bands-added-image}.
Fig. \ref{fig:features-compare} shows that trials in both 2-valence and 4-emotion classes can be separated to a better degree (although not optimal) when using the VGG features from the EEG-PSD combined image than directly using the EEG-PSD features. The EEG-PSD features only form distinct clusters for each subject and are unable to separate the valence/emotion classes whereas the VGG features allow for better separation.

\subsection{Evaluating individual modality performance}
This section presents the classification results obtained by using individual modalities on the four datasets. Table \ref{table-singular-modality} shows accuracy and mean F1-score results for individual modalities.

\begin{table}[!ht]
\centering
\caption{Individual Modality Performance Evaluation}
\label{table-singular-modality}
{%
\begin{tabular}{|p{.4in}|p{.4in}p{.4in}p{.4in}p{.4in}p{.4in}|}
\hline
\textbf{Response} & \textbf{EEG} & \textbf{Cardiac} & \textbf{GSR} & \textbf{Face-1} & \textbf{Face-2}
\\\hline
\multicolumn{6}{|c|}{\textbf{DEAP Dataset}} \\ \hline
\textbf{Valence} & 71.09/0.68 & 70.86/0.69 & 70.70/0.68 & 71.08/0.68 & 72.28/0.70 \\
\textbf{Arousal} & 72.58/0.65 & 71.09/0.63 & 71.64/0.65 & 72.21/0.65 & 74.47/0.68 \\
\textbf{Liking} & 74.77/0.65 & 74.77/0.64 & 75.23/0.64 & 75.60/0.62 & 76.69/0.62 \\
\textbf{Emotion} & 48.83/0.26 & 45.55/0.31 & 45.94/0.25 & 43.52/0.28 & 46.27/0.27 \\ \hline
\multicolumn{6}{|c|}{\textbf{AMIGOS Dataset}} \\ \hline
\textbf{Valence} & 83.02/0.80 & 81.89/0.80 & 80.63/0.79 & 80.58/0.77 & 77.28/0.74 \\
\textbf{Arousal} & 79.13/0.74 & 82.74/0.76 & 80.94/0.74 & 83.10/0.76 & 77.28/0.72 \\
\textbf{Liking} & 85.27/0.81 & 82.53/0.77 & 80.47/0.72 & 80.27/0.72 & 79.81/0.72 \\
\textbf{Emotion} & 55.71/0.30 & 58.08/0.36 & 56.41/0.34 & 57.74/0.28 & 56.79/0.27 \\ \hline
\multicolumn{6}{|c|}{\textbf{MAHNOB-HCI Dataset}} \\ \hline
\textbf{Valence} & 80.77/0.76 & 78.76/0.73 & 78.98/0.73 & 83.04/0.79 & 85.13/0.82 \\
\textbf{Arousal} & 80.42/0.72 & 78.76/0.74 & 81.84/0.75 & 82.15/0.77 & 81.57/0.76 \\
\textbf{Emotion} & 57.86/0.33 & 57.23/0.35 & 57.84/0.32 & 60.41/0.35 & 63.42/0.35 \\ \hline
\multicolumn{6}{|c|}{\textbf{DREAMER Dataset}} \\ \hline
\textbf{Valence} & 78.99/0.75 & 80.43/0.78 & --- & --- & --- \\
\textbf{Arousal} & 79.23/0.77 & 80.68/0.77 & --- & --- & --- \\
\textbf{Emotion} & 54.83/0.33 & 57.73/0.36 & --- & --- & ---
\\\hline
\end{tabular}%
}\bigskip

\normaljustify
Cardiac features refer to features extracted using PPG in the DEAP dataset and using ECG in the AMIGOS, MAHNOB-HCI, and DREAMER datasets. Face-1 and Face-2 refer to the methods \rmnum{3}D.1 and \rmnum{3}D.2 respectively. Valence, Arousal, and Liking have been classified into two classes (50\% chance accuracy) whereas Emotion has been classified into four classes (25\% chance accuracy). All values denote the mean percentage accuracy followed by the mean F1-score (separated by ``/") whereas missing values represent missing modality data.
\end{table}

It is clear from Table \ref{table-singular-modality} that our results in multiple categories for all the four datasets are better than those reported previously, as shown in Table \ref{table-DEAP}. The CNN based features that we extracted for all modalities contribute most towards this classification improvement for all the modalities. Furthermore, for all four datasets and for all modalities, the performance is substantially greater than the chance accuracy. EEG proves to be the best performing bio-sensing modality whereas Cardiac and GSR features also perform very well despite containing fewer channels. Furthermore, the frontal-video-based showed high accuracy in the affective classification for the three datasets and surpassed the accuracy obtained by the bio-sensing modalities in many cases.

The classification performance using various modalities even for varying trial length is consistently better than that reported in previous studies (Table \ref{table-DEAP}). Our results surpass the previous best results obtained by using only individual modalities for the DEAP and MAHNOB-HCI datasets and the baseline accuracies for the AMIGOS and DREAMER datasets. Furthermore, we find higher accuracy for Liking classification than for Valence/Arousal for DEAP and AMIGOS datasets, suggesting that it might be easier for subjects to rate their likeness for the video contents than rating valence and arousal. This is understandable since the latter terms are difficult to comprehend than Liking and depend highly on the physiological baseline of the subject at any particular time.

\subsection{Evaluating multi-modality performance}
This section presents the results of combining different modalities for affective state classification. Specifically, as shown in Table \ref{table-multi-modality}, we first combine the three bio-sensing modalities (only two for the DREAMER since it does not contain GSR data) to evaluate their joint performance and then the EEG and Face-video modalities through the CNN-VGG-extracted features. Finally, for the DEAP dataset, we present the results of training an LSTM network with the time-varying features from the EEG and Face-video modalities (see Section \rmnum{3}.E). 

\begin{table}[!ht]
\centering
\caption{Multi-modality Performance Evaluation}
\label{table-multi-modality}
{%
\begin{tabular}{|p{0.45in}|p{0.45in}p{0.45in}p{0.45in}p{0.45in}|}
\hline
\textbf{Response} & \textbf{Bio-sensing} & \textbf{EEG and Face} & \textbf{EEG and Face (LSTM)} & \textbf{Previous Best Accuracy}
\\\hline
\multicolumn{5}{|c|}{\textbf{DEAP Dataset}} \\ \hline
\textbf{Valence} & 71.87/0.68 & 73.94/0.69 & 79.52/0.70 & 77.19 \\
\textbf{Arousal} & 73.05/0.68 & 74.13/0.66 & 78.34/0.69 & 76.17 \\
\textbf{Liking} & 75.86/0.69 & 76.74/0.63 & 80.95/0.70 & 68.40 \\
\textbf{Emotion} & 49.53/0.27 & 48.11/0.28 & 54.22/0.31 & 50.80 \\ \hline
\multicolumn{5}{|c|}{\textbf{AMIGOS Dataset}} \\ \hline
\textbf{Valence} & 83.94/0.82 & 78.23/0.74 & --- & ---  \\
\textbf{Arousal} & 82.76/0.76 & 81.47/0.72 & --- & --- \\
\textbf{Liking} & 83.53/0.77 & 81.49/0.75 & ---  & --- \\
\textbf{Emotion} & 58.56/0.40 & 58.02/0.29 & --- & --- \\ \hline
\multicolumn{5}{|c|}{\textbf{MAHNOB-HCI Dataset}} \\ \hline
\textbf{Valence} & 80.36/0.75 & 85.49/0.82 & --- & 73.00 \\
\textbf{Arousal} & 80.61/0.71 & 82.93/0.77 & --- & 68.50 \\
\textbf{Emotion} & 58.07/0.30 & 62.07/0.35 & --- & --- \\ \hline
\multicolumn{5}{|c|}{\textbf{DREAMER Dataset}} \\ \hline
\textbf{Valence} & 79.95/0.77 & --- & --- & 62.49  \\
\textbf{Arousal} & 79.95/0.77 & --- & --- & 62.32 \\
\textbf{Emotion} & 55.56/0.33 & --- & --- & ---
\\\hline
\end{tabular}%
}\bigskip

\normaljustify
Bio-sensing refers to combining features from EEG, ECG/PPG, and GSR signals. 
\newline EEG + Face refers to combining features from EEG- and video-based modalities. 
\newline EEG + Face (LSTM) refers to combining features from EEG- and video-based modalities for every second in the trial to train an LSTM model.
Due to the trial length varying widely in the AMIGOS and MAHNOB-HCI datasets, the LSTM-based method could not be applied to them. The DREAMER dataset does not have video data.
\end{table}

\begin{figure}[!ht] \centering
{\includegraphics[width=3.45in]{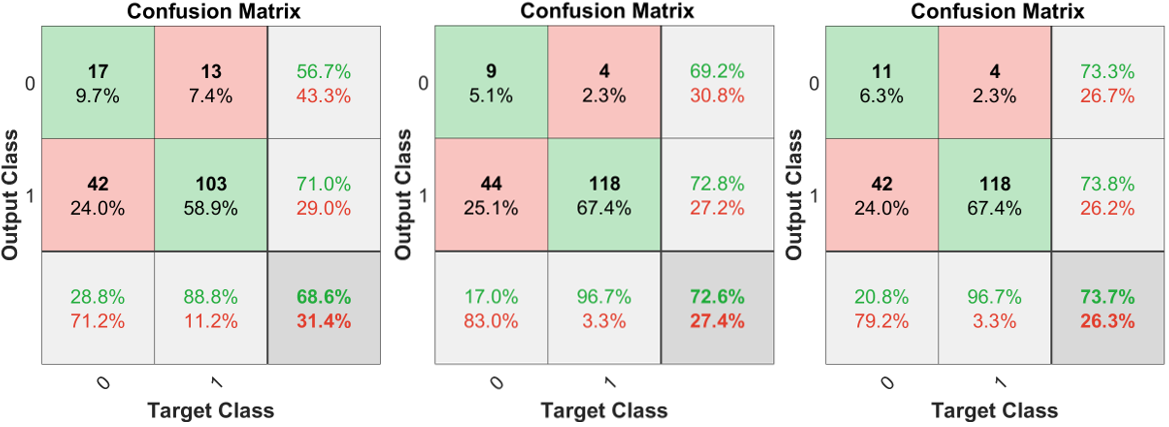}}
\vspace{-15pt}
\caption{Confusion matrix for DEAP Dataset valence using EEG alone (left), Faces alone (middle), and EEG + Faces (right) for a sample 80/20 dataset distribution.}
\label{fig:confusion-matrix}
\end{figure}

In almost all cases, we find that combining features from multiple modalities increases classification accuracy. The fusion of features from bio-sensing modalities increases the accuracy in many cases for all the four datasets. We also note that by training the LSTM network with the features from EEG and Face-video modalities not only increases the accuracy as compared to the individual modalities (from Table \ref{table-singular-modality}) but also outperform the best accuracy on the DEAP dataset reported in Table \ref{table-DEAP}. For AMIGOS, MAHNOB-HCI, and DREAMER datasets, we see that using multiple modalities outperform single-modality accuracy in many cases and sets up new benchmarks by beating previous best results. As an example of the increased performance by combining multiple modalities over single modalities, we also show a confusion matrix of the classification in Fig. \ref{fig:confusion-matrix}. We also performed two-sample t-Test between Bio-Sensing and EEG plus Face multi-modal combinations for the datasets. The p-values of the t-Test analysis for the valence, arousal, liking, and emotion classification for the DEAP  dataset were 0.676, 0.543, 0.939 and 0.347 respectively. The p-values for the valence, arousal, liking, and emotion classification for the AMIGOS dataset were 0.003, 0.266, 0.134 and 0.026 respectively. The p-values for the valence, arousal, and emotion classification for the MAHNOB-HCI Dataset were 0.0134, 0.293 and 0.149. We could not perform similar t-Test on the DREAMER dataset since it does not contain Video (Face) modality data.

\begin{table}[!ht]
\centering
\caption{Combined Dataset Performance Evaluation}
\label{table-combined-dataset}
\begin{tabular}{|p{.4in}|p{.4in}p{.4in}p{.4in}p{.4in}p{.4in}|}
\hline
\textbf{Response} & \textbf{EEG} & \textbf{Cardiac} & \textbf{GSR}  & \textbf{Face-1} & \textbf{Face-2}
\\\hline
\multicolumn{6}{|c|}{\textbf{DEAP + AMIGOS Combined Dataset}} \\ \hline
\textbf{Valence} & 62.80/0.58 & 59.69/0.59 & 59.64/0.58 & 63.04/0.62 & 62.38/0.62 \\
\textbf{Arousal} & 62.27/0.61 & 63.61/0.61 & 61.98/0.62 & 67.66/0.65 & 68.65/0.66 \\
\textbf{Liking} & 69.13/0.59 & 69.27/0.61 & 69.27/0.55 & 67.99/0.64 & 68.65/0.64 \\
\textbf{Emotion} & 37.47/0.27 & 37.50/0.22 & 37.24/0.31 & 40.92/0.36 & 42.24/0.36 \\ \hline
\multicolumn{6}{|c|}{\textbf{DEAP + AMIGOS + MAHNOB-HCI Combined Dataset}} \\ \hline
\textbf{Valence} & 61.24/0.60 & 58.57/0.59 & 58.98/0.57 & 61.59/0.61 & 62.56/0.63 \\
\textbf{Arousal} & 65.15/0.63 & 61.84/0.61 & 61.02/0.59 & 65.94/0.65 & 67.15/0.66 \\
\textbf{Emotion} & 40.21/0.35 & 36.33/0.31 & 35.71/0.28 & 42.51/0.33 & 43.00/0.32
\\\hline
\end{tabular} 
\bigskip

\normaljustify
The DEAP + AMIGOS combined dataset consists of the data from 72 subjects and more than 1,900 trials. The DEAP + AMIGOS + MAHNOB-HCI combined dataset consists of the data from 99 subjects and more than 2,400 trials. Only the deep-learning-based methods are used for extracting features for evaluation from various modalities because these can be extracted from all datasets in the same manner.
\end{table}

\begin{table}[!ht]
\centering
\caption{Transfer Learning Performance Evaluation}
\label{table-transfer-dataset}
\begin{tabular}{|p{.4in}|p{.4in}p{.4in}p{.4in}p{.4in}p{.4in}|}
\hline
\textbf{Response} & \textbf{EEG} & \textbf{Cardiac} & \textbf{GSR}  & \textbf{Face-1} & \textbf{Face-2}
\\\hline
\multicolumn{6}{|c|}{\textbf{DEAP + AMIGOS (Train Dataset), MAHNOB-HCI (Test Dataset)}} \\ \hline
\textbf{Valence} & 63.55/0.60 & 64.77/0.54 & 64.96/0.55 & 55.02/0.52 & 62.01/0.62 \\
\textbf{Arousal} & 58.37/0.55 & 62.50/0.52 & 62.50/0.52 & 59.32/0.54 & 58.60/0.58 \\
\textbf{Emotion} & 36.65/0.32 & 39.58/0.28 & 38.64/0.28 & 36.38/0.39 & 34.05/0.37  \\ \hline
\multicolumn{6}{|c|}{\textbf{DEAP (Train Dataset), MAHNOB-HCI (Test Dataset)}} \\ \hline
\textbf{Valence} & 62.70/0.54 & 63.59/0.46 & 65.19/0.47 & 56.48/0.49 & 59.86/0.59 \\
\textbf{Arousal} & 61.99/0.55 & 61.46/0.48 & 63.23/0.52 & 59.33/0.56 & 61.99/0.60 \\
\textbf{Emotion} & 35.88/0.23 & 38.01/0.24 & 39.08/0.24 & 33.57/0.33 & 32.50/0.22
\\\hline
\end{tabular} 
\bigskip

\normaljustify
Only the deep-learning-based methods are used for extracting features for evaluation from various modalities because these can be extracted from all datasets in the same manner.
\end{table}

\subsection{Evaluating the classification performance using combining datasets and transfer learning}
To show that the proposed deep-learning-based features are independent of the number of EEG channels, trial length, the image resolution of the video, ECG/PPG cardiac modality, etc., we trained the ELM classifier with data from more than one datasets. We also use a transfer-learning approach to train the ELM classifier with data from some of the four datasets and then test it against the remaining dataset. The combined datasets were randomly divided into an 80:20 ratio for training and testing. This allows us to verify how scalable our feature extraction methods across datasets having discrepancies in recording devices (e.g. ECG vs PPG) and parameters (e.g. channel numbers). Table \ref{table-combined-dataset} shows that despite all these discrepancies across the datasets, our methods work well and always perform considerably better than the chance accuracy and the baseline accuracies for individual datasets \cite{DEAP,AMIGOS,MAHNOB-HCI} reported in Table \ref{table-DEAP}. Table \ref{table-transfer-dataset} shows the results of training with two datasets and testing on the third. The above combinations of datasets were chosen because all the sensor modalities were used in the datasets and the DEAP dataset contains more trials (1,280 trials) than the other two datasets combined together (AMIGOS and MAHNOB-HCI containing 640 and 540 trials respectively). Even when we test the ELM on a dataset, the trials from which were not used for training, the results were consistently better (more so for ECG/PPG and GSR modalities) than many previous studies and far above the chance accuracy. The slight decrease in performance for some modalities compared to those trained with the data from the same dataset might be due to two factors, namely, the varying trial length between the datasets and only using the VGG-based features common to the datasets (for consistency among the datasets) as opposed to combining features from other methods like conditional entropy, HRV, face-localization, etc.

\begin{figure}[!ht] \centering
{\includegraphics[width=3.45in]{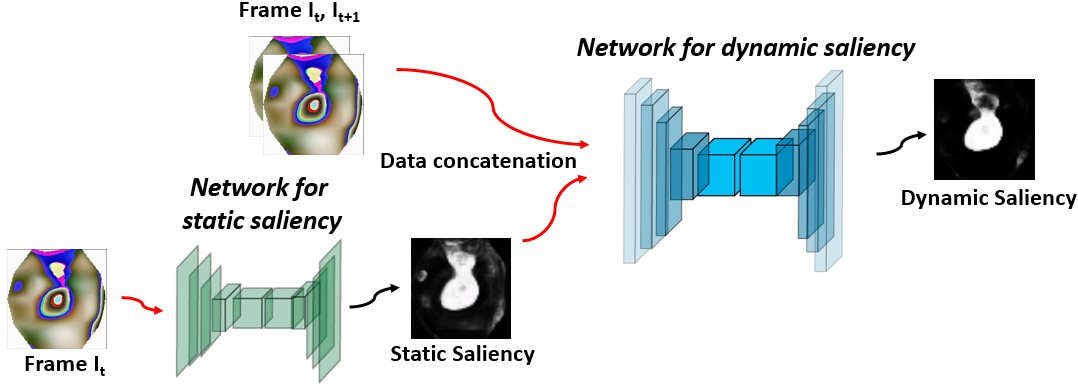}}
\caption{Salient brain regions corresponding to low/high valence/arousal in DEAP dataset. The frontal lobe has high activation.}
\label{fig:SaliencyNetwork}
\end{figure}

\subsection{Identifying the salient brain regions that contribute towards processing various emotions}
As is clear from the performance evaluation sections above, the deep-learning-based methods are able to extract more meaningful features and perform better than traditional features. This section aims to explore what insights the proposed deep-learning-based method can provide on the brain regions contributing to emotional responses. To this end, we added a reverse VGG network (before the final max pooling step) to the pre-trained VGG network that extracted the informative features we used above. That is, we added a deconvolving network to the convolving network. As shown in \cite{Salient_Object_Detection}, the convolution-deconvolution network can be used to identify the most salient areas in the images in both static and dynamic manner. We utilize this network (Fig. \ref{fig:SaliencyNetwork}) to detect those regions in the EEG-PSD brain images that contribute most towards processing various emotions. The pairs of EEG-PSD images for consecutive seconds for a trial $I_{t}, I_{t+1}$ were sent to the dynamic convolution-deconvolution network along with the output of the static saliency for the image at $I_{t}$. The static saliency network identified the most salient areas whereas the dynamic saliency network was able to learn the variations between these image areas for every consecutive second. This procedure was done for every second for all the trials. We report results only from the DEAP dataset for this method because of its fixed trial length.

\begin{figure}[!ht] \centering
{\includegraphics[width=3.5in]{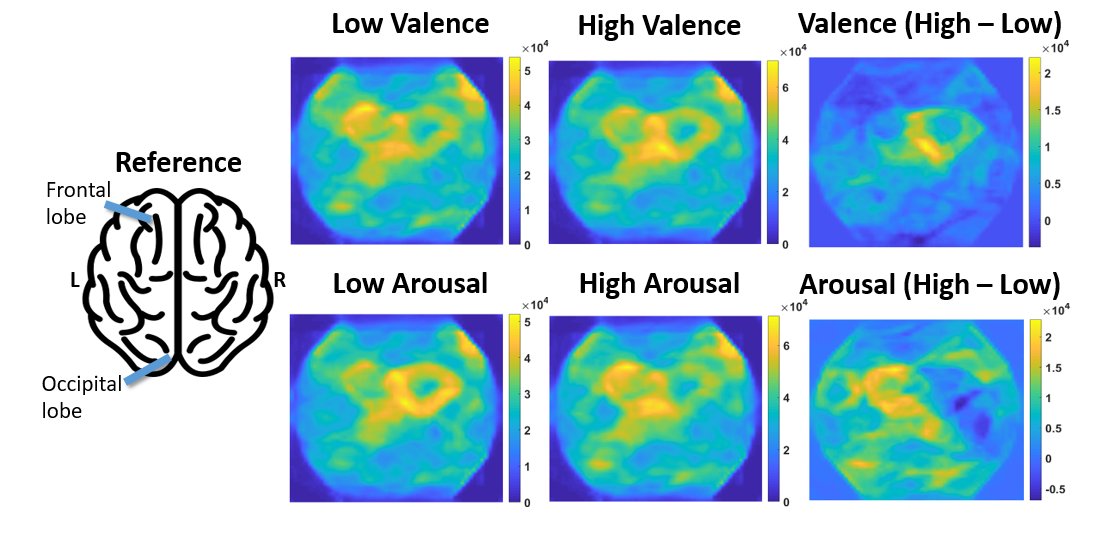}}
\caption{Convolution-Deconvolution network on EEG-PSD images to identify salient brain regions corresponding to affective states. Pixels in individual images were scaled between 0 and 1.}
\label{fig:brain-salient-regions}
\end{figure}

We used the RGB combined images (Figure \ref{fig:bands-added-image}) for every second for every trial of the low/high valence/arousal instances from the DEAP dataset by first convolving and then deconvolving them in the network described above. Hence, theoretically, the areas with most salient variations across the trials would represent the brain regions that are most receptive to the particular affective state. Fig. \ref{fig:brain-salient-regions} shows the brain activity for these affective states after averaging the output of the network across all the trials for the affective state (valence/arousal). Most of the activity is over the frontal and central lobes around the FC3, FCz, FC4, and Cz locations according to the EEG 10-20 system. This is consistent with the textbook evidence regarding the processing of human emotions \cite{frontal_lobe_1,frontal_lobe_2}. More interestingly, we observe from the difference image between high and low arousal that the processing of arousal affective state is much more widely distributed across the brain than valence. Hence, this method allows us to use a single image to represent such areas across the brain, and across all subjects and trials, that are most activated for a particular affective measure rather than using multiple such EEG images. We present these results as a starting point to take this work towards using the EEG for investigating the generation and processing of emotions inside the brain. 

\section{Discussion and Conclusion}
Advances in deep-learning have not translated into bio-sensing and multi-modal affective computing research domain mostly due to the absence of very-large-scale datasets. Such datasets are available for vision/audio modalities due to the ease of data collection. Hence, for the time being, it seems that the only viable solution is to use ``off-the-shelf" pre-trained deep-learning networks to extract features from bio-sensing modalities. The proposed methods present the advantages of being scalable and able to extract features from different datasets. Such ``off-the-shelf" features prove to work better than the traditionally used features of various bio-sensing modalities.

This study proposed novel methods to affective computing research by employing deep-learning features across various modalities. We showed that these methods perform better than previously reported results on four different datasets containing various recording discrepancies. The methods were also evaluated on the combined datasets. Furthermore, the various modalities were fused to augment the performance of our models. The LSTM was used to learn the temporal dynamics of the features during stimulus presentation and increase the classification accuracy, compared to averaging the features across that trial. We also showed that features extracted from bio-sensing modalities such as EEG can be combined with those from the video-based modality to increase the accuracy further. In the future, we will investigate the elicitation of emotions and its dependence on the physiological baseline. We also plan to work on ``real-world'' emotion recognition problems where the subjects are mobile while responding to audio-visual stimuli present in the environment as opposed to being displayed on a screen in a well-controlled laboratory.

\section*{Acknowledgment}
The authors would like to thank the research groups collecting and disseminating datasets used in this research and for granting us access to the datasets. This work was supported in part by the Army Research Laboratory under Cooperative Agreement Number W911NF-10-2-0022, NSF NCS-1734883, NSF 1540943, and a seed grant from UC San Diego Center for Wearable Sensors.

\ifCLASSOPTIONcaptionsoff
  \newpage
\fi

\begin{IEEEbiography}[{\includegraphics[width=1in,height=1.25in,clip]{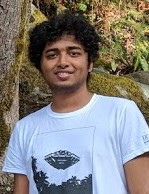}}]{Siddharth}(S'14) received B.Tech. in Electronics and Communications Engineering at the Indian Institute of Information Technology, Allahabad, India in 2015 and M.S. in Intelligent Systems, Robotics, and Control at the University of California San Diego (UCSD), La Jolla, USA in 2017. Currently, he is pursuing Ph.D. at UCSD where his research interests include computational neuroscience, multi-modal bio-sensing, and affective computing. He has also undertaken research internships at National University of Singapore, French National Center for Scientific Research, Samsung Research America, and Facebook Reality Labs.
\end{IEEEbiography}

\begin{IEEEbiography}[{\includegraphics[width=1in,height=1.25in,clip,keepaspectratio]{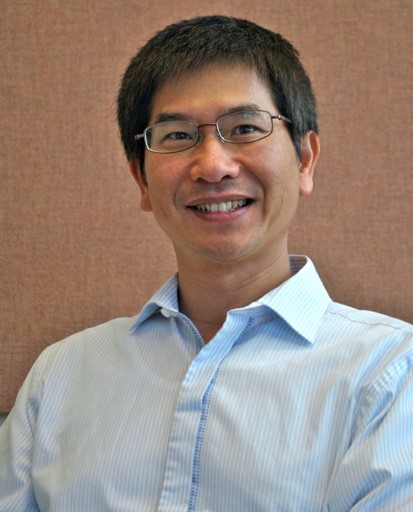}}]{Tzyy-Ping Jung} (S'91-M'92-SM'06-F'15) received the B.S. degree in electronics engineering from National Chiao Tung University, Hsinchu, Taiwan, in 1984, and the M.S. and Ph.D. degrees in electrical engineering from The Ohio State University, Columbus, OH, USA, in 1989 and 1993, respectively. He is an Associate Director of the Swartz Center for Computational Neuroscience, Institute for Neural Computation, and an Adjunct Professor of Bioengineering at UCSD. His research interests are in the areas of biomedical signal processing, cognitive neuroscience, machine learning, human EEG, functional neuroimaging, and brain-computer interfaces and interactions.
\end{IEEEbiography}

\begin{IEEEbiography}[{\includegraphics[width=1in,height=1.25in,clip,keepaspectratio]{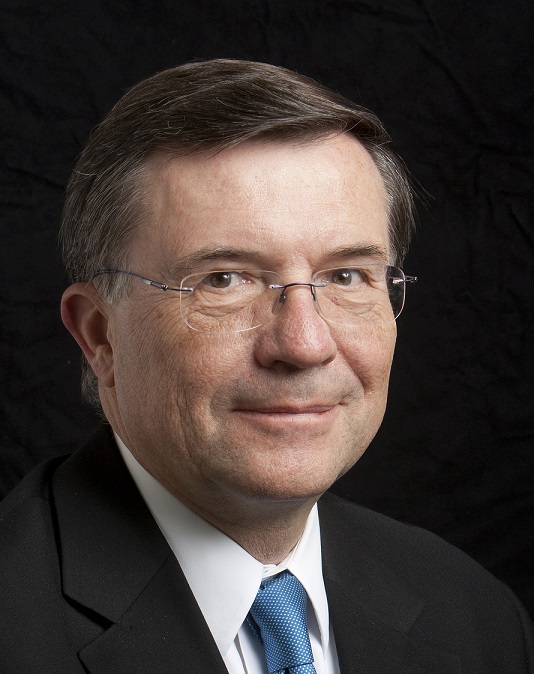}}]{Terrence J. Sejnowski} (F'00) is the Francis Crick Professor at The Salk Institute for Biological Studies, where he directs the Computational Neurobiology Laboratory, and a Professor of Biology and Computer Science and Engineering at the University of California San Diego, where he is co-Director of the Institute for Neural Computation. The long-range goal of Dr. Sejnowski's research is to understand the computational resources of brains and to build linking principles from brain to behavior using computational models. Dr. Sejnowski has published over 500 scientific papers and 12 books, including The Computational Brain, with Patricia Churchland. He received the Wright Prize for Interdisciplinary research in 1996, the Hebb Prize from the International Neural Network Society in 1999, and the IEEE Neural Network Pioneer Award in 2002.  He is a member of the National Academy of Sciences, the National Academy of Engineering, the National Academy of Medicine and the National Academy of Inventors.
\end{IEEEbiography}

\end{document}